\documentclass[11pt]{article}
\usepackage{coling2018}
\usepackage{times}
\usepackage{url}
\usepackage{latexsym}
\usepackage{graphicx}
\usepackage{url}
\usepackage{multirow}
\usepackage{amsmath}
\usepackage{amsfonts,amssymb}
\usepackage{subfigure}
\title{Modeling Multi-turn Conversation with Deep Utterance Aggregation}

\author{Zhuosheng Zhang$^{1,2,}$\thanks{$\ $ These authors contribute equally. $\dagger$ Corresponding author. This paper was partially supported by
		National Key Research and Development Program of China (No. 2017YFB0304100),
		National Natural Science Foundation of China (No. 61672343 and No. 61733011),
		Key Project of National Society Science Foundation of China (No. 15-ZDA041),
		The Art and Science Interdisciplinary Funds of Shanghai Jiao Tong University (No. 14JCRZ04).} , Jiangtong Li$^{1,2,3,*}$, Pengfei Zhu$^{1,2,5}$, Hai Zhao$^{1,2,\dagger}$, Gongshen Liu$^{4}$\\
	$^{1}$Department of Computer Science and Engineering, Shanghai Jiao Tong University \\
	$^{2}$Key Laboratory of Shanghai Education Commission for Intelligent Interaction \\ and Cognitive Engineering, Shanghai Jiao Tong University, Shanghai, 200240, China\\
	$^{3}$College of Zhiyuan, Shanghai Jiao Tong University, China\\
	$^{4}$School of Cyber Security, Shanghai Jiao Tong University, China\\
	$^{5}$School of Computer Science and Software Engineering, East China Normal University, China\\
	{\tt \{zhangzs, keep\_moving-lee\}@sjtu.edu.cn, 10152510190@stu.ecnu.edu.cn,} \\ {\tt zhaohai@cs.sjtu.edu.cn, lgsheng@sjtu.edu.cn}
}

\date{}

\begin{document}
\maketitle
\begin{abstract}

Multi-turn conversation understanding is a major challenge for building intelligent dialogue systems. This work focuses on retrieval-based response matching for multi-turn conversation whose related work simply concatenates the conversation utterances, ignoring the interactions among previous utterances for context modeling. In this paper, we formulate previous utterances into context using a proposed deep utterance aggregation model to form a fine-grained context representation. In detail, a self-matching attention is first introduced to route the vital information in each utterance. Then the model matches a response with each refined utterance and the final matching score is obtained after attentive turns aggregation. Experimental results show our model outperforms the state-of-the-art methods on three multi-turn conversation benchmarks, including a newly introduced e-commerce dialogue corpus.

\end{abstract}

\section{Introduction}

\blfootnote{
	%
	%
	%
	%
	%
	%
	This work is licensed under a Creative Commons 
	Attribution 4.0 International License.
	 License details:
	\url{http://creativecommons.org/licenses/by/4.0/}
}

Human-computer interactive systems are booming due to their promising potentials and alluring commercial values \cite{Qiu2017AliMe,Cui2017SuperAgent,Zhao17,Huang2018Moon,Jia2014A}. With the development of neural models \cite{Zhang2018Neural,He2018Syntax,li2018Seq,Cai2018Seq,zhang2018OneShot}, building an intelligent dialogue system as our personal assistant or chat companion, is no longer a fantasy, among which multi-turn natural language understanding still keeps extremely challenging, requiring the system to comprehend the conversation context and reply in an informative and coincident manner.

Multi-turn conversation modeling plays a key role in dialogue systems, either for generation-based \cite{iulian17,iulian17-2,ganbin17,Wu2017Neural} or retrieval-based ones \cite{Wu2016Sequential,Zhou2016Multi} in which the latter is the focus of this paper. A natural approach for multi-turn modeling is simply concatenating the context utterances \cite{Lowe2015The,yan2016Learning}. However, this will introduce much noise since previous utterances as the context is lengthy and redundant. The gist is to identify pertinent information in previous utterances and properly model the utterance relationships to ensure conversation consistency. To avoid unnecessary information loss, \cite{Wu2016Sequential} matches a response with each utterance in the context, paying little attention on distrinct importance of each utterance and also failing to touch internal semantics inside utterances.

\begin{figure*}
	\centering
	\includegraphics[width=1.0\textwidth]{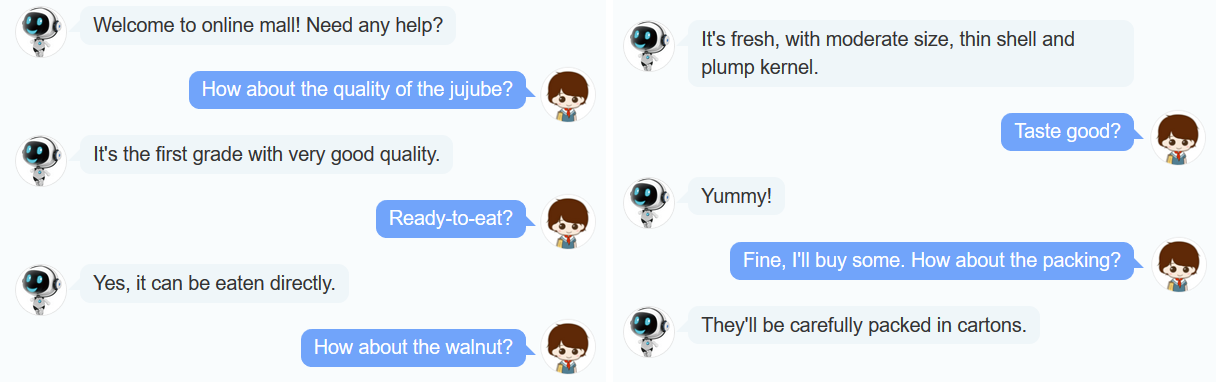}
	\caption{\label{tab:example} An example of E-commerce Dialogue Corpus.}
\end{figure*}

In fact, the relevance of each utterance to the supposed response usually varies. As shown in Figure \ref{tab:example}, the last utterance in a conversation empirically conveys the user intention while the other utterances depict the conversation in different aspects \footnote{For a multi-turn conversation, we define the latest  user utterance (or called current message) as the \emph{last utterance}, which is waiting for a response.}. Thus, instead of considering all the conversation turns equally, we have to weigh previous conversations in a more sophisticated way. With a turns-aware aggregation design, our model alleviates the drawback of previous work.

In addition, words in an utterance also hold different importance to the whole utterance representation. Our solution is to employ attention-based recurrent networks on each utterance against utterance itself, aggregating the vital pieces of the contextual utterances.

Finally, in conjunction with this paper, we release an E-commerce Dialogue Corpus (ECD) to facilitate the related studies. To our best knowledge, this is the first public e-commerce dataset for dialogue system development that is extracted from real human conversations. Different from previous datasets that only focus on a single type of dialogue like chitchat, this dataset is more comprehensive due to diverse types of conversations (e.g. \emph{commodity consultation, logistics express, recommendation, negotiation} and \emph{chitchat}) concerning various commodities. Our improved retrieval-based multi-turn dialogue response matching model is evaluated on three benchmark datasets, including our newly released one, giving state-of-the-art performance.

The rest of this paper is organized as follows. The next section reviews related work. Our proposed model is introduced in Section 3, then the experiments and analysis are reported in Section 4, followed by the conclusion in Section 5.

\section{Related Work}
With the impressive success of various referential  natural  language
processing studies \cite{Zhang2016Probabilistic,Cai2017Pair,zhang2018NHD,Qin2017Adversarial,zhang2018SubMRC,Bai2018deep}, developing an intelligent dialogue system becomes realizable, which means training machines to converse with human in natural languages \cite{williams-asadi-zweig,he-EtAl,dhingra-EtAl,Zhang2018Personalizing}. Towards this end, a number of data-driven dialogue systems are designed \cite{Lowe2015The,Wu2016Sequential,wenN2N17,hongyuan17,Young2017Augmenting,Lipton2016BBQ}, in which modeling multi-turn conversation has drawn more and more attention. To acquire a contextual response, previous utterances are taken as input. \newcite{Lowe2015The} concatenated  all previous utterances and last utterance as the context representation and then computed the matching degree score based on the context representation to encode candidate response. \newcite{yan2016Learning} selected the previous utterances in different strategies and combined them with last utterance to form a reformulated context. \newcite{Zhou2016Multi} performed context-response matching with a multi-view model on both word level and utterance level. \newcite{Wu2016Sequential} improved the leveraging of utterances relationship and contextual information by matching a response with each utterance in the context based on a convolutional neural network.

Different from previous studies, our model for the first time discriminates the importance of previous conversations and also accumulates substantial parts from each utterance according to each word in the utterance itself in a multi-turn scenario.

\section{Deep Utterance Aggregating Strategy}
\begin{figure*}
	\centering
	\includegraphics[width=1\textwidth]{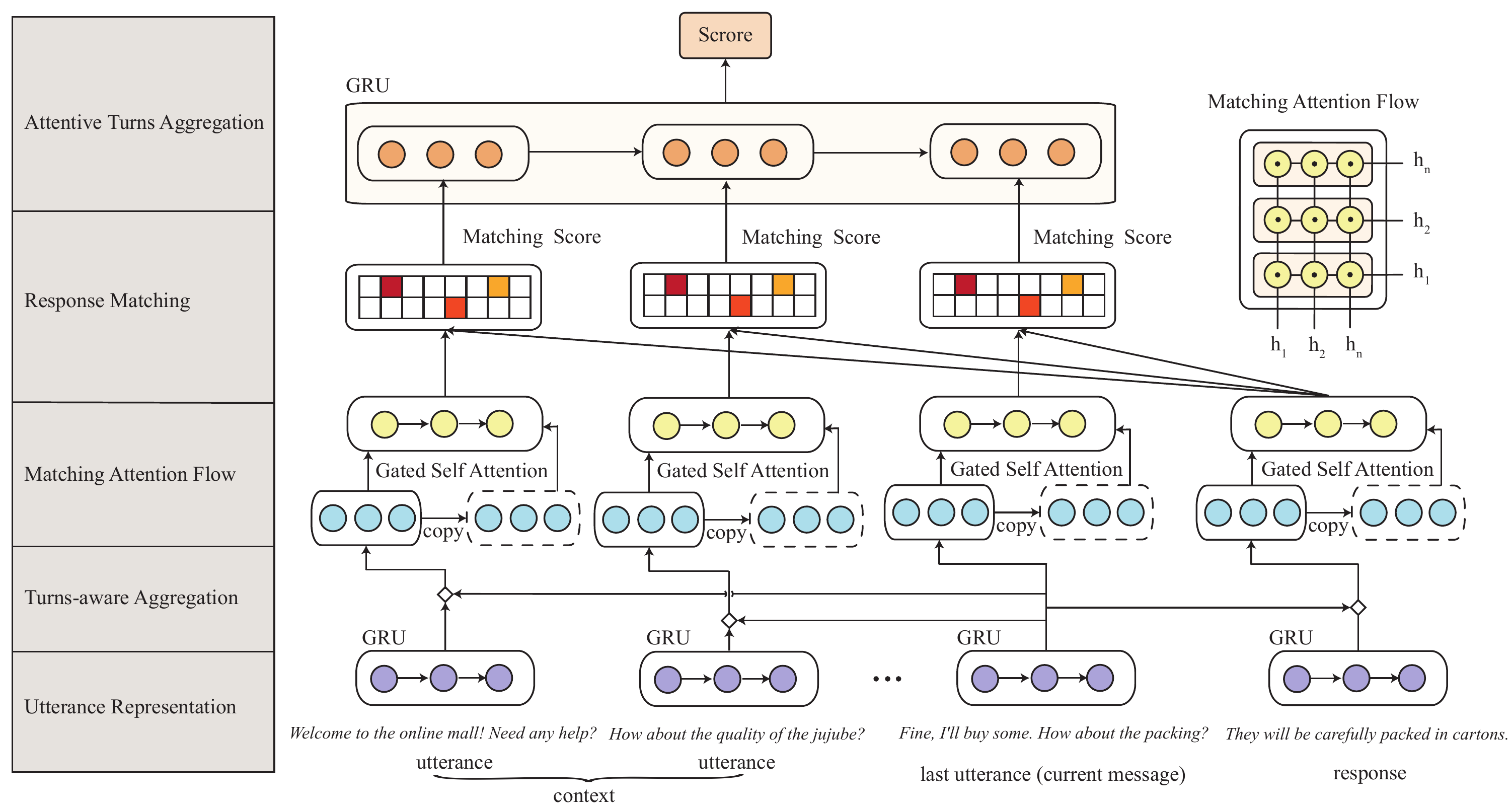}
	\caption{Structure overview of the proposed dialogue system.}
	\label{fig:overview}
\end{figure*}
Each conversation in the concerned multi-turn response retrieval task can be described as a triple $<C,R,Y>$. $ C = \{U_{1}, ... , U_{t}\} $ is the conversation context where $ \{U_{k}\} $ denotes the $k$-th utterance. $ R $ is a response of the conversation and $ Y $ belongs to $ \{0,1\} $, where $ Y_i = 1 $ means the response is proper, otherwise $ Y_i = 0 $. The aim is to build a discriminator $ \mathcal{F(\cdot,\cdot)} $ on $<C,R,Y>$. For each context-response pair $ \{C, R\} $, $ \mathcal{F}(C, R) $ measures the matching score of the pair.

In this section, we will introduce our Deep Utterance Aggregation (DUA) model for the multi-turn conversation task. Figure \ref{fig:overview} shows the architecture. DUA formulizes utterances into the context and mines the key information from utterances and response. Then DUA conducts semantic matching between each utterance and the response candidate to obtain a matching score. Specifically, there are five modules within DUA. Each utterance or response is fed to the first module to form an utterance or response embedding. The second module combines the last utterance with the preceding utterances. Then, the third module filters the redundant information and mines the salient feature within the utterances and response. The fourth module matches the response and each utterance at both word and utterance levels to feed a Convolutional Neural Network (CNN) for encoding into matching vectors. In the last module, the matching vectors are delivered to a gated recurrent unit (GRU) \cite{Cho2014Learning} in chronological order of the utterances in the context and the final matching score of $ \{U, R\} $ is obtained. 

DUA is superior to existing models in the following ways. First, the last utterance which is the most important in dialogue is especially fused within preceding utterances, thus the key guideline information from the last utterance can be handled in a more semantically pertinent way. Second, in each utterance, the salient information can be highlighted and those redundant pieces will be neglected to some extent, both of which can effectively guide the later response matching. Third, after attentive turns aggregation, the connections in the conversation are accumulated again to calculate the matching scores.

\subsection{Utterance Representation}
To use deep neural networks, symbolic data needs to be transformed into distributed representations, namely, word embedding \cite{bengio2003neural,mikolov:2013}.
Given a context-response pair, $ \{C, R\} $ whose context are split into utterances, $ C = \{U_{1} , ... , U_{t}\} $, a lookup table is used to map each word into a low-dimensional vector. Let $n_{u}$ and $n_{r}$ denote the length of the $ k $-th utterance and response, $ U_k $ and $ R $ can be represented as $ U_{k} = [u_{1}, ... , u_{n_{u}}] $ and $ R = [r_{1}, ... , r_{n_{r}}] $, where $ u_{i}, r_{i}$ are the $ i $-th word in the utterance and response respectively. 

To encode each utterance and response, we employ a GRU to propagate information along the word sequence of $U_k$ and $R$. Suppose $H_k = [h_1, ... , h_n]$ is the hidden states of the input sequence, the structure of GRU is described as follows.
\begin{equation}
\begin{split}
&z_i = \sigma(W_zu_i + V_zh_{i-1})  \\
&r_i = \sigma(W_ru_i + V_rh_{i-1}) \\
&\tilde{h_{i}} = tanh(W_hu_i + V_h (r_i \odot h_{i-1})) \\
&h_{i} = z_i \odot \tilde{h_{i}} + (1-z_i) \odot h_{i-1}
\end{split}
\end{equation}
where $ \sigma{(\cdot)} $ is the sigmoid function, $z_{i}$ and $r_{i}$ are the update and reset gates respectively, $\odot$ denotes the element-wise multiplication, and $ W_z, W_r, W_h, V_z, V_r, V_h $ are parameters. We fed each utterance and response sequence to the GRUs and obtain the utterance representation $S_k$ and response representation $S_{r}$, respectively.

\subsection{Turns-aware Aggregation}
Encoding the utterance sequence and response in the above way, there comes a drawback that all the utterances in the conversation are fairly dealt with, which fails to mine the connections between the last utterance and the rest preceding utterances. Thus, a first-stage turns-aware aggregation mechanism is proposed to address this problem.

Let $S = [S_{1}, ... , S_{t}, S_{r}]$ denote the representation of the utterances and response. Suppose $ F = [F_{1}, ... ,F_{t}, F_{r}] $ is the fusion of each $S_j \in S$ with the last utterance $S_t$, for each $ \forall j \in \{1, ... , r\}$, $ F_{j} \in $ $ F $, we define the fusion of the utterance as
\begin{equation}
\begin{split}
F_{j} = S_{j} \diamond S_{t}
\end{split}
\end{equation}
where $\diamond$ denotes the aggregation operation. In this work, we adopt a simple concatenation strategy\footnote{We empirically investigated \emph{concatenation}, \emph{element-wise summation}, \emph{element-wise multiplication} in this work and $concatenation$ strategy shows the best performance.}. So far, the turns-aware representation $F$ is obtained via aggregation. 
\subsection{Matching Attention Flow}
After turns-aware aggregation, the representations of the preceding utterances and response have been refined by the last utterance. However, the sequences are quite lengthy and redundant, which makes it hard to distill the pivotal information. In order to address this problem, we adopt a self-matching attention mechanism to directly match the fused representation against itself, which is similar as that adopted in  \cite{Wang2017Gated}. It dynamically collects information from the input sequence and filters the redundant information. 
Suppose $ \hat{F} = [f_{1}, ... , f_{n}] \in F$  is the input and $P = [p_{1}, ... , p_{n}]$ is the output of the self-matching attention on response, then $\forall t$, $ p_{t} \in P$ is defined as

\begin{equation}
p_{t} = GRU(p_{t-1}, [f_{t}, c_{t}])
\end{equation}
where $GRU(\cdot,\cdot)$ denotes the same calculation as $(1)$, $ [\cdot,\cdot] $ is the concatenation of two vectors and $ c_{t} = att(\hat{F}, f_{t})$ is the result of the self-matching attention. $ \forall t $, $ f_{t} \in \hat{F}$, $ c_{t}$ is defined as 
\begin{equation}
\begin{split}
&s_{j}^t = v^Ttanh(W_{v}f_{j} + W_{\tilde{v}}f_{t} + b_r) \\
&a_{i}^t = exp(s_{i}^t) / \sum_{j=1}^{n}exp(s_{j}^t) \\
&c_{t} = \sum_{i=1}^{n}a_{i}^tf_{i}
\end{split}
\end{equation}
where $ W_{r} $, $ W_{\tilde{v}} $, $b_r $ are the parameters and $v^T$ is a context matrix which is randomly initialized and jointly trained.

Self-matching attention pinpoints important parts from the utterance according to the current word and the whole utterance representation through fusing each previous utterance and the last utterance.

\subsection{Response Matching}

Following \cite{Wu2016Sequential}, we use word-level and utterance-level representations to build two matching matrices and employ CNN to obtain salient matching information from the matrices. Suppose we have matching matrices $M_1$ and $M_2$ in word-level and utterance-level for each utterance-response pair. Then, $ \forall k$, $ U_{k} \in U$ and $ \forall (i,j) $, the $ (i,j) $-th element of $M_1$ and $M_2$ is defined respectively

\begin{equation}
e_{1, i, j} = u_{i}^T r_{j} \\
\end{equation}
\begin{equation}
e_{2, i, j} = P_{u_i}^T A P_{r_j}
\end{equation}
where $P_{u_i}$ and $P_{r_j}$ denote the outputs of the utterance and response after \emph{Matching Attention Flow} respectively. $A \in$ $\mathbb{R}^{c\times c}$ is a linear transforming matrix.

A convolutional operation followed by a max-pooling operation will be applied to $M_1$ and $M_2$ for each utterance. The convolutional layer is used to extract and combine local features from adjacent words and the following max-pooling layer forms the representations for the current word. For the convolutional operation, a group of filter matrices $K$ with variable sizes $l * l$ and bias $b$ are utilized. The filter transforms the word matrices $M_1$ and $M_2$ to another two matrices $M_{1c}$ and $M_{2c}$. $\forall i$ $k \in (1,2) $, the transformed matrices $M_{kc}$ is define as:

\begin{equation}
 	M_{{kc}, [i][j]} = ReLU(\sum_{i}^{i+l-1}\sum_{j}^{j+l-1}K \cdot M_{{k,[i:i+l-1][j:j+l-1]}} + b)
\end{equation}
where $i$ and $j$ index the row $i$-th and column $j$-th element, respectively. Next, a max-pooling operation is adopted and the representation ${m}_p$ for $p$-th utterance in a conversation is obtained through flattening and  concatenating the two matrices after pooling as follows:

\begin{equation}
	\hat{m}_{k, [i][j]} = \max(M_{kc,[i:i+l-1][j:j+l-1]}) 
\end{equation}
\begin{equation}
	m_p = [flatten(\hat{m}_1) \oplus flatten(\hat{m}_2)]
\end{equation}
where $flatten()$ is the flatten operation and $\oplus$ is the concatenation operation.

\subsection{Attentive Turns Aggregation}
To aggregate the matching information of the attentive turns in the last stage, The outputs of CNN, $M=[m_1, ... ,m_{n}]$ are fed to GRU to obtain $ H_m = [h_{m_1}, ... ,h_{m_n}] $. $\forall i$, $h_{m,i} \in H_m$ is defined as 

\begin{equation}
h_{m,i} = GRU(h_{m_{i-1}}, m_i)
\end{equation} 
where $GRU(\cdot,\cdot)$ denotes the same calculation and parameterization as Eq.$(1)$. Suppose $v_f = L(H_m)$ is the attention operation which is defined as:
\begin{equation}
\begin{split}
&t_i = v^T\tanh(W_{t}P_{u_i} + V_{t}h_{m_i} + b) \\
&\alpha_i = \exp(t_i)/\sum_{j=1}^{n}\exp(t_j) \\
&L(H_m) = \sum_{i=1}^{n}\alpha_ih_{m_i} 
\end{split}
\end{equation}
where $W_{t}$, $V_{t}$ and $b$  are parameters. With $v_f$, we define $ \mathcal{F}(U, R)$ as:
\begin{equation}
\mathcal{F}(U, R) = softmax(W_s v_f)
\end{equation}
where $W_s$ is the parameter. During the training phase, model parameters are updated according to a cross-entropy loss.

Note that \emph{Turns-aware Aggregation} and \emph{Attentive Turns Aggregation} can be seen as two stages of interaction across the utterances (we call all these two process as ``\emph{Context Fusion}" henceforth). Specifically, the former is simply a combination after the \emph{Utterance Representation} for richer turns-aware information while the latter is to  aggregate matching states of previous turns after attention learning against each utterance itself and the response.

\section{Experiment}

\subsection{Dataset}

\begin{table*}
	\centering
	{
		\begin{tabular}{cccccccccc}
			\hline
			
			\hline
			& \multicolumn{3}{c}{Ubuntu} & \multicolumn{3}{c}{Douban}  & \multicolumn{3}{c}{ECD}  \\
			&Train & Valid & Test & Train & Valid & Test & Train & Valid & Test  \\
			\hline
			\# context-response pairs & 1M & 500K & 500K &1M & 50K & 10K  &1M & 10K & 10K\\
			\# candidates per context  & 2 & 10  & 10 & 2 & 2 & 10& 2 & 2 & 10\\
			Avg \#  turns per context & 10.13 & 10.11 & 10.11 & 6.69 & 6.75 & 6.45 & 5.51 & 5.48 & 5.64\\
			Avg \# words per utterance & 11.35 & 11.34 & 11.37 & 18.56 & 18.50 & 20.74 & 7.02 & 6.99 & 7.11\\
			\hline
			
			\hline
		\end{tabular}
	}
	\caption{\label{tab:dataset}  Data statistics template for latter use.}
\end{table*}

We evaluate our model on three multi-turn conversation datasets, the Ubuntu Dialogue Corpus (Ubuntu) \cite{Lowe2015The}, the Douban Conversation Corpus (Douban) \cite{Wu2016Sequential} and our released E-commerce Dialogue Corpus (ECD) \footnote{Our released dataset  along with source code can be accessed via \url{https://github.com/cooelf/DeepUtteranceAggregation}.}. Data statistics are in Table \ref{tab:dataset}. 

\paragraph{Ubuntu Dialogue Corpus}
\emph{Ubuntu Dialogue Corpus} consists of multi-turn human-computer conversations constructed from Ubuntu IRC chat logs. The training set contains 1 million label-context-response triples where the original context and corresponding response are labeled as positive and negative response are selected randomly on the dataset. On both validation and test sets, each context contains one positive response and 9 negative responses. 
\paragraph{Douban Conversation Corpus}
\emph{Douban conversation corpus} is an open domain dataset constructed from Douban group which is a popular social networking service in China. Response candidates on the test set are collected by a standard search engine \emph{Apache Lucene}\footnote{\url{http://lucene.apache.org/}}, other than negative sampling without human judgment on Ubuntu Dialogue Corpus. That is, the last turn of each Douban dialogue with additional keywords extracted from the context on the test set is used as query to retrieve 10 response candidates from the Lucene index set. 

\paragraph{E-commerce Dialogue Corpus} In this part, we will introduce our E-commerce Dialogue Corpus. Though previously described public datasets have served in solid studies, there is no comprehensive e-commerce dataset available for research. We collect real-world conversations between customers and customer service staff from our E-commerce partners in Taobao \footnote{\url{https://www.taobao.com}}, which is the largest e-commerce platform in China \footnote{All the data have been carefully desensitized and anonymized with the consent of our partners and avoid privacy issues.}. It contains over 5 types of conversations (e.g. commodity consultation, logistics express, recommendation, negotiation and chitchat) based on over 20 commodities. As word segmentation treatment is the primary step in Chinese language processing tasks \cite{Zhao2007A,Cai2017Fast,Cai2016Neural}, we adopt \emph{BaseSeg} \cite{Hai2006An} to tokenize the texts. For a discriminative learning, we add negative responses by ranking the response corpus based on the last utterance along with the top-5 key words in the context using \emph{Apache Lucene}. The ratio of the positive and the negative is 1:1 in training and validation, and 1:9 in testing. 
\subsection{Settings}

Our evaluation is based on the following information retrieval metrics: Mean Average Precision (MAP), Mean Reciprocal Rank (MRR), Precision at 1 (P@1) and  Recall at position $k$ in $n$ candidates ($R_{}n@k$) , which are widely used for relevance evaluation \cite{Wu2016Sequential,Lowe2015The}. For the sake of computational efficiency, the maximum number of utterances is specialized as 10 and each utterance contains at most 50 words. We apply truncating and zero-padding when necessary. Word embedding is trained by Word2Vector \cite{mikolov:2013} on the training data and the dimension is 200. Our model is implemented using the Theano \footnote{\url{https://github.com/Theano/Theano}}. We use stochastic gradient descent with ADAM \cite{Kingma2014Adam} updates for optimization. The batch size is 200 and the initial learning rate is 0.001. The window size of convolution and pooling is (3, 3) and the number of hidden units for the character GRU is set to 200. All of our models are run on a single GPU (GeForce GTX 1080 Ti). We run all the models up to 5 epochs and select the model that achieves the best result in validation. 

Our baselines include:

$\bullet$ \textbf{Single-turn matching models}: Basic models in \cite{Kadlec2015Improved,Lowe2015The}, including TF-IDF, CNN, RNN, LSTM and biLSTM ; We also explore other advanced single-turn matching models, MV-LSTM \cite{Wan2016Match}, Match-LSTM \cite{Wang2015Learning},
Attentive-LSTM \cite{Tan2015LSTM}, Multi-Channels \cite{Wu2016Sequential}; These models concatenate the context utterances together to match a response.

$\bullet$ \textbf{Advanced multi-turn matching models}:
Multi-view model of \cite{Zhou2016Multi} that models utterance relationships from word sequence view and utterance sequence view; Deep Learning-to-Respond (DL2R) model of \cite{yan2016Learning} which reformulates the last utterance (query) with other utterances in the context via neural model; Sequential Matching Network (SMN) \cite{Wu2016Sequential} that matches a response with each utterance in the context.

The results of baseline models on Ubuntu and Douban are from \cite{Wu2016Sequential}. For evaluation on our ECD dataset, we reproduce the models following their same settings.

\begin{table*}[t]\centering
	
	{
		\begin{tabular}{l|l|l|l|l|l|l|l|l|l}
			\hline
			
			\hline
			\multirow{2}{*}{Model}& \multicolumn{3}{c|}{Ubuntu Dialogue Corpus} & \multicolumn{6}{c}{Douban Conversation Corpus} \\
			\cline{2-10}
			&${\rm R_{10} @1} $& ${\rm R_{10} @2} $ & ${\rm R_{10} @5} $ & MAP  &  MRR  & P@1 &${\rm R_{10} @1} $& ${\rm R_{10} @2} $ & ${\rm R_{10} @5} $ \\
			\hline
			TF-IDF & 0.410 & 0.545 & 0.708 & 0.331 & 0.359 & 0.180 & 0.096 & 0.172 & 0.405 \\
			RNN &0.403&0.547&0.819&0.390&0.422&0.208&0.118&0.223&0.589 \\
			CNN &0.549&0.684&0.896&0.417&0.440&0.226&0.121&0.252&0.647 \\
			LSTM &0.638&0.784&0.949&0.485&0.537&0.320&0.187&0.343&0.720 \\
			BiLSTM &0.630&0.780&0.944&0.479&0.514&0.313&0.184&0.330&0.716 \\
			\hline
			Multi-View &0.662&0.801&0.951&0.505&0.543&0.342&0.202&0.350&0.729 \\
			DL2R &0.626&0.783&0.944&0.488&0.527&0.330&0.193&0.342&0.705 \\
			MV-LSTM &0.653&0.804&0.946&0.498&0.538&0.348&0.202&0.351&0.710 \\
			Match-LSTM &0.653&0.799&0.944&0.500&0.537&0.345&0.202&0.348&0.720 \\
			Attentive-LSTM &0.633&0.789&0.943&0.495&0.523&0.331&0.192&0.328&0.718 \\
			Multi-Channel &0.656&0.809&0.942&0.506&0.543&0.349&0.203&0.351&0.709 \\
			Multi-Channel$_{exp}$ &0.368&0.497&0.745&0.476&0.515&0.317&0.179&0.335&0.691 \\
			SMN &0.726&0.847&0.961&0.529&0.569&0.397&0.233&0.396&0.724 \\
			\hline
			DUA &\textbf{0.752}&\textbf{0.868}&\textbf{0.962}&\textbf{0.551}&\textbf{0.599}&\textbf{0.421}&\textbf{0.243}&\textbf{0.421}&\textbf{0.780} \\
			\hline
			
			\hline
		\end{tabular}
		
	}
	\caption{\label{tab:result} Comparison of different models on Ubuntu Dialogue Corpus and Douban Conversation Corpus. All the results except ours are from \cite{Wu2016Sequential}.}
\end{table*}

\begin{table}[t]
	\centering
	{
		\begin{tabular}{l|l|l|l}
			\hline
			
			\hline		
			Model & ${\rm R_{10} @1} $& ${\rm R_{10} @2} $ & ${\rm R_{10} @5} $ \\
			\hline
			TF-IDF  & 0.159 & 0.256 & 0.477   \\
			RNN  & 0.325 & 0.463  & 0.775 \\
			CNN & 0.328 & 0.515  & 0.792 \\
			LSTM  & 0.365 & 0.536 & 0.828  \\
			BiLSTM  & 0.355 & 0.525 & 0.825  \\
			\hline
			Multi-View & 0.421 & 0.601 & 0.861 \\
			DL2R & 0.399 & 0.571 & 0.842 \\
			MV-LSTM  & 0.412 & 0.591  &  0.857  \\
			Match-LSTM  & 0.410 & 0.590  & 0.858 \\
			Attentive-LSTM  & 0.401 & 0.581 & 0.849\\
			Multi-Channel & 0.422 & 0.609 & 0.871 \\
			Multi-Channel$_{exp}$  & 0.352 & 0.556 & 0.827 \\
			SMN  & 0.453 & 0.654 & 0.886 \\
			\hline
			DUA & \textbf{0.501} & \textbf{0.700} & \textbf{0.921} \\
			\hline
			
			\hline
		\end{tabular}
		
	}
	\caption{\label{tab:result_osc} Comparison of different models on E-commerce Dialogue Corpus.}
\end{table}

\subsection{Experimental Results}
Table \ref{tab:result}-\ref{tab:result_osc} show the results on the three corpora. Our model outperforms all other models greatly in terms of most of the metrics. Single matching models which concatenate the previous utterances, perform much worse than our model, showing the importance of utterance relationships and simply concatenating utterances together is not an appropriate solution for multi-turn conversation modeling. Our model also achieves a great improvement (4.8\% ${\rm R_{10} @1}$ on ECD corpus) over the state-of-the-art multi-turn response matching model, SMN, which matches each utterance and response without turns-aware aggregation and matching attention flow. This comparison indicates the effectiveness of our context composing approach. The advantage on ECD dataset further indicates our model can well imitate the conversations of real customer service instead of merely being good at chitchat.

\subsection{Discussion}

\paragraph{Conversation Type Analysis}
To evaluate the model performance on different types of conversations, we manually separate our ECD test set into 5 categories.

\textbf{$\bullet$ Consultation}: consultations about commodity's property, usage, packaging, etc. 

\textbf{$\bullet$ Logistics}: questions about logistics partners, delivery progress. 

\textbf{$\bullet$ Recommendation}: commodity comparisons and recommendations.

\textbf{$\bullet$ Negotiation}: customer complaints and negotiations.

\textbf{$\bullet$ Chitchat}: greetings, non task-oriented conversations and chitchats.

Table \ref{tab:type_ECD} shows the statistics and the model results. As we see, the types of chitchat and logistics tend to be easily handled. Recommendations, consultation and negotiations are relatively harder to respond since they often involve with various topics (e.g. the concerned commodities) and intentions, which makes our corpus more challenging than previous chitchat or question answering based corpora. 

\begin{figure*}\centering
	
	\subfigure[Highlighted utterance]{
		\begin{minipage}[b]{0.445\textwidth}
			\includegraphics[width=1\textwidth]{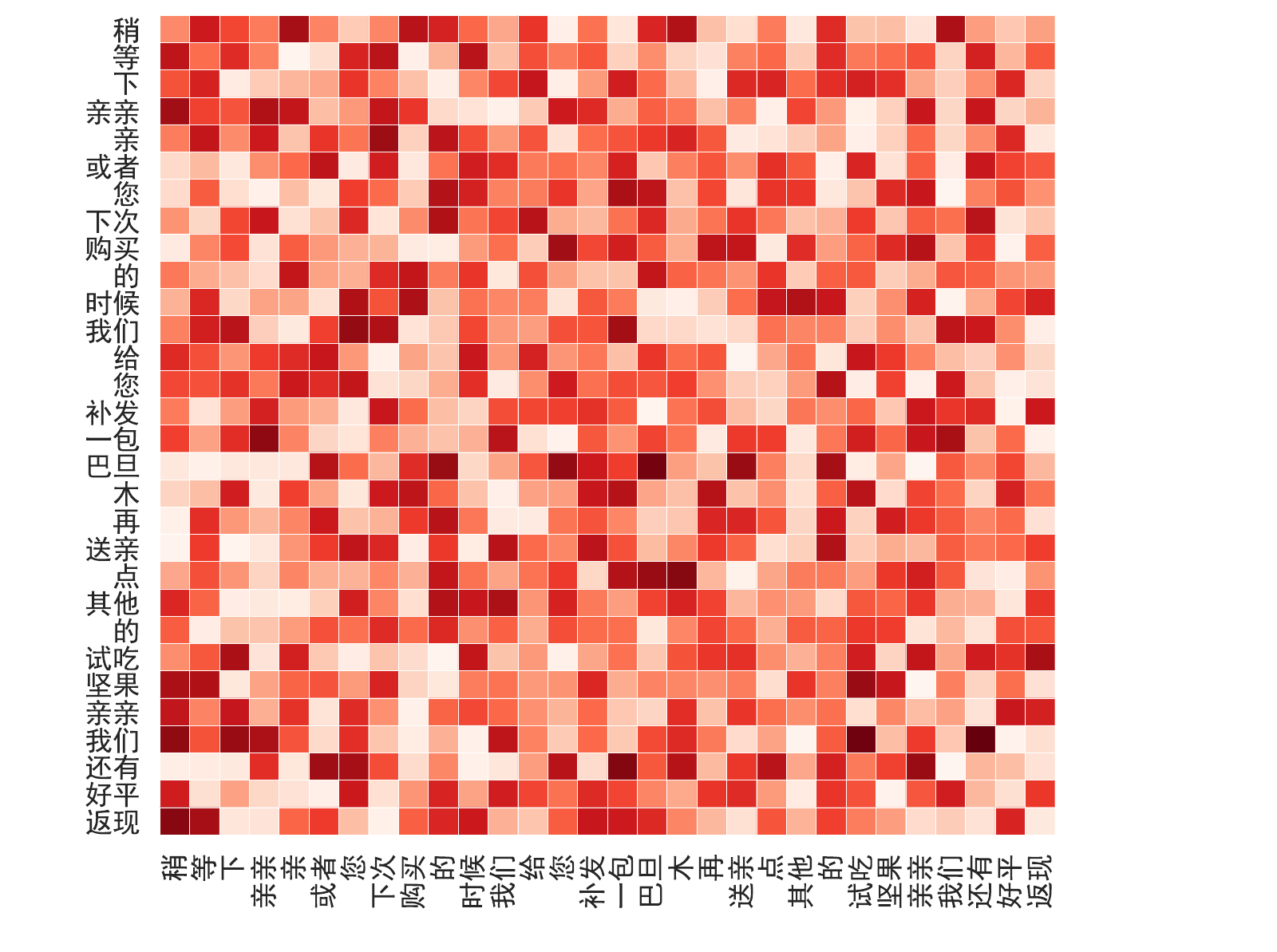}
		\end{minipage}
	}
	\subfigure[Response]{
		\begin{minipage}[b]{0.445\textwidth}
			\includegraphics[width=1\textwidth]{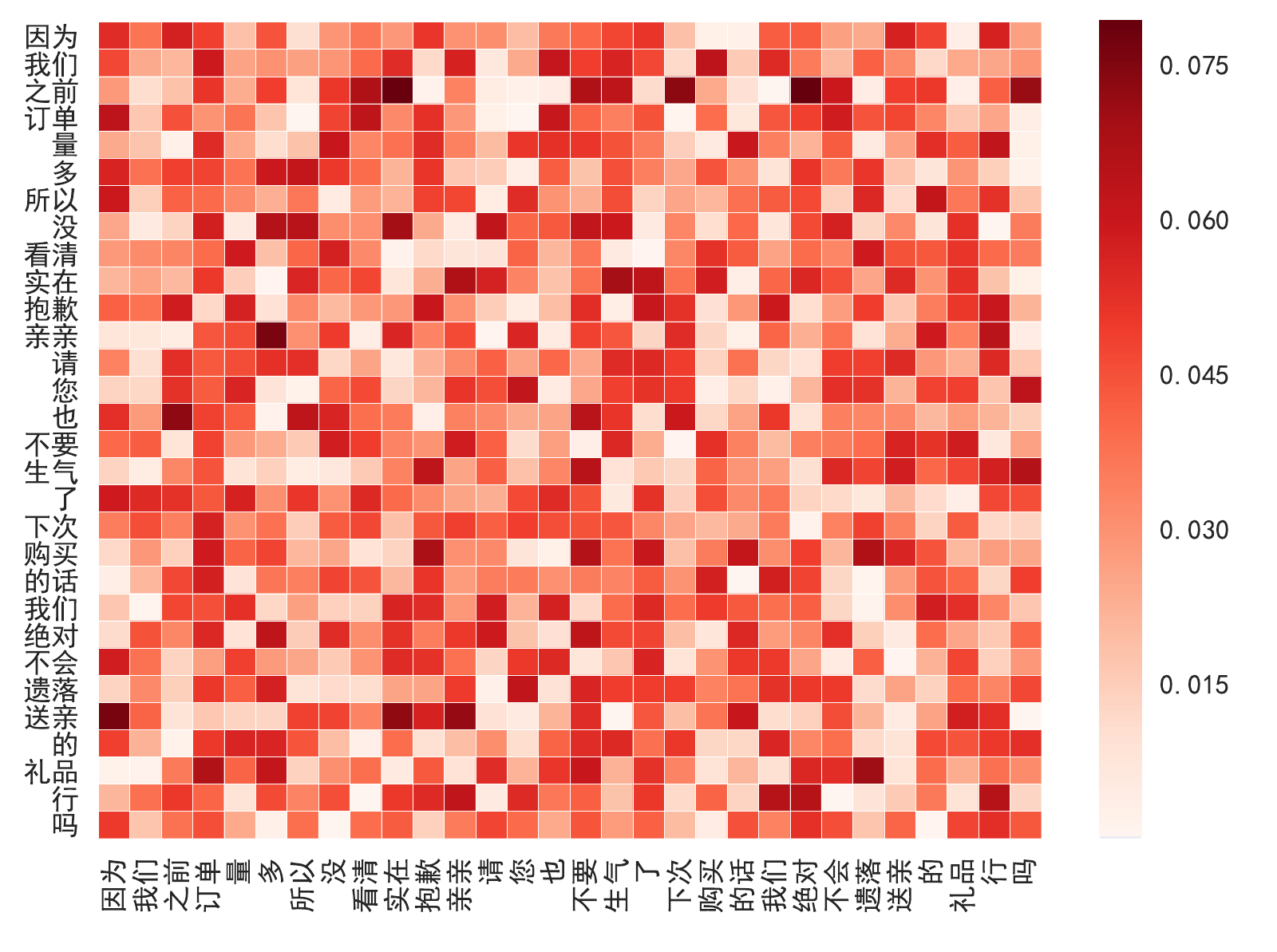}
		\end{minipage}
	}
	\\\begin{flushleft}
	\scriptsize{\emph{
			Last utterance (user): How can you miss my gift! And the delivery is so slow. You are spoiled !!!
			\\Highlighted utterance (bot): please wait a moment dear. For compensation, we'll reissue you a bag of almonds at your next consumption. Besides, we will also send you some nuts to taste. If you give us five-star rating and comments, you'll also receive some cashback.     
			\\Response: Because we had too many orders before, we unfortunately misread your order. We are really sorry for the mistake. Please don't be angry, we will never forget your gift again.} }\end{flushleft}
	
	\caption{Pair-wise attention visualization on utterance and response after matching attention flow.} \label{fig:attention}	
\end{figure*}

\paragraph{Visualization}
To analyze the effectiveness of the attention mechanism of our model, we draw the self-matching distributions after \emph{matching attention flow}. From the validation set of our ECD data, Figure \ref{fig:attention} shows the word weights of a momentous utterance (with high weights in the response matching component) and the response respectively. We see the model could accurately distill the linchpin from the utterance, \emph{\{Next consumption, reissue, a bag of almond, send you, some nuts, cashback\}} and from the response \emph{\{too many orders before, really sorry, don't be angry, your gift\}}. When a user complained about the missing gift and slow delivery, our model could distinguish the user's intention after self-matching and seek out the suitable response substantially according to the crux of the presented utterance. This shows our model is effective at selecting the vital points after \emph{Matching Attention Flow}, guiding the \emph{Response Matching} layer to collect more relevant pieces.

\begin{table}[t]
	\centering
	{
		\begin{tabular}{l|l|l|l}
			\hline
			
			\hline
			& ${\rm R_{10} @1} $ & ${\rm R_{10} @2} $ & ${\rm R_{10} @5} $  \\
			\hline
			Consultation (36.1\%) & 0.474 & 0.696  & 0.900 \\
			Logistics （(7.3\%) & 0.510 & 0.707 & 0.916   \\
			Recommendation (4.4\%) & 0.487 & 0.590 & 0.897    \\
			Negotiation (5.9\%) & 0.385 & 0.462 & 0.846    \\
			Chitchat (26.3\%) & 0.573 & 0.762 & 0.931 \\
			\hline
			Overall(100\%) & 0.501 & 0.700 & 0.921 \\
			\hline
			
			\hline
		\end{tabular}
		
	}
	\caption{\label{tab:type_ECD} Results on different types of conversations.}
\end{table}

\begin{table}[t]
	\centering
	{
		\begin{tabular}{l|l|l|l}
			\hline
			
			\hline
			&${\rm R_{10} @1} $& ${\rm R_{10} @2} $ & ${\rm R_{10} @5} $ \\
			\hline
			DUA & 0.501 & 0.700  & 0.921   \\
			-CF & 0.453 & 0.642 & 0.890    \\
			-MAF  & 0.432 & 0.625 & 0.883 \\
			-CF -MAF & 0.413 & 0.613 & 0.867     \\
			\hline
			
			\hline
		\end{tabular}
		
	}
	\caption{\label{tab:abs_ecd} Ablation study on ECD dataset. CF and MAF denote the \emph{Context Fusion} and \emph{Matching Attention Flow}. The bracket means the context fusion approach adopted by the model.}
\end{table}

\paragraph{Ablation Study}

To have an insight of the effectiveness of each component in DUA, we remove one each time. The steepest reduction (6.9\% ${\rm R_{10} @1}$) is observed when we remove \emph{Matching Attention Flow} which shows it quite vital to draw the linchpins of each utterance. The performance also drops substantially (4.8\% ${\rm R_{10} @1}$) when removing \emph{Context Fusion} including the first turns-aware aggregation (first-stage aggregation) and replacing the last GRU (last-stage aggregation) for matching accumulation with a multi-layer perceptron. This indicates that utterance relationships are indeed important. Without \emph{Context Fusion} and \emph{Matching Attention Flow} mechanisms, the model performs the worst which verifies our proposed mechanism indeed improves the context representation essentially.

\subsection{Error Analysis}
After carefully analyzing the predicted responses, we find the error cases could be classified into the following categories for later further improvement.

\paragraph{Multiple intentions}
In E-commerce conversations users extremely likely express various intentions in a single message, which is another big difference from previous multi-turn conversation corpus besides diverse types of conversations among various commodities. For example, \emph{\{User: How about the packaging of skin care products. By the way, which delivery company will be responsible for shipping and how long can I receive the goods?\}}. This would seriously confuse the model where the given response might be preferential to one or another aspect.

\paragraph{Topic errors}
Our model retrieves response according to semantic similarity with the context, with no special attention on the conversation topic, such as the currently discussed commodities. In most cases, the concerned commodity would be picked out from the context with high attention weights and guide the model to select responses. However, when the conversation involves several goods, for example, \emph{\{User: How about nuts? Bot: Nuts is good. User: Ok then, how about zongzi?\}}, the model might give the response about \emph{nuts} instead of \emph{zongzi}. This indicates there exists much potential for improvements by considering extra topic recognition.

\paragraph{Multiple suitable responses}
In our ECD dataset, we assume there is only one correct response for each conversation which is the same setting as Ubuntu Dialogue Corpus. However, the model sometimes gives responses having similar meaning with the ground-truth one, but they would be regarded as wrong during evaluation, especially for fairly long conversations. This could make the task rather challenging with the strict restriction of exact match. This  might be alleviated by involving expert labeling like \cite{Wu2016Sequential}. However, this is quite labour-intensive and subjective. In the future, we would explore more automatic solutions.

\section{Conclusion}
In this paper, we propose a deep utterance aggregation approach to form a fine-grained context representation. We also release the first e-commerce dialogue corpus to research communities. Experiments on three datasets show the model can yield new state-of-the-art results. Various analyses are conducted to evaluate the model and the released dataset. In the future, we may study how to improve modeling of contextual semantics and design a better neural network for multi-turn conversations in terms of various intentions and topics.

\bibliography{acl2018}
\bibliographystyle{acl}

\end{document}